\documentclass[letterpaper]{article} % DO NOT CHANGE THIS
\usepackage{aaai2026}  % DO NOT CHANGE THIS
\usepackage{times}  % DO NOT CHANGE THIS
\usepackage{helvet}  % DO NOT CHANGE THIS
\usepackage{courier}  % DO NOT CHANGE THIS
\usepackage[hyphens]{url}  % DO NOT CHANGE THIS
\usepackage{graphicx} % DO NOT CHANGE THIS
\usepackage{natbib}  % DO NOT CHANGE THIS AND DO NOT ADD ANY OPTIONS TO IT
\usepackage{caption} % DO NOT CHANGE THIS AND DO NOT ADD ANY OPTIONS TO IT
\usepackage{algorithm}
\usepackage{algorithmic}

\usepackage{booktabs}
\usepackage{multirow}
\usepackage{amsmath}
\usepackage{xspace}
\usepackage{xcolor}
\usepackage{colortbl}
\usepackage{utfsym}
\usepackage{amssymb}
\usepackage{pifont}
\usepackage{makecell}
\usepackage{tcolorbox}

\usepackage{newfloat}
\usepackage{listings}
\DeclareCaptionStyle{ruled}{labelfont=normalfont,labelsep=colon,strut=off} % DO NOT CHANGE THIS
\floatstyle{ruled}
\newfloat{listing}{tb}{lst}{}
\floatname{listing}{Listing}
\title{Beyond the Visible: Benchmarking Occlusion Perception in Multimodal\\ Large Language Models}
\author{
    Zhaochen Liu\textsuperscript{\rm 1,2},
    Kaiwen Gao\textsuperscript{\rm 3}\equalcontrib,
    Shuyi Liang\textsuperscript{\rm 3}\equalcontrib,
    Bin Xiao\textsuperscript{\rm 3}\equalcontrib,\\
    Limeng Qiao\textsuperscript{\rm 4},
    Lin Ma\textsuperscript{\rm 4},
    Tingting Jiang\textsuperscript{\rm 1,5}\thanks{Corresponding author.}
}
\affiliations{
    \textsuperscript{\rm 1}National Engineering Research Center of Visual Technology, National Key Laboratory\\ for Multimedia Information Processing, School of Computer Science, Peking University\\
    \textsuperscript{\rm 2}AI Innovation Center, School of Computer Science, Peking University\\
    \textsuperscript{\rm 3}School of Electronics Engineering and Computer Science, Peking University\\
    \textsuperscript{\rm 4}Meituan Inc.
    \textsuperscript{\rm 5}National Biomedical Imaging Center, Peking University\\
}

\usepackage{bibentry}
\begin{document}

\maketitle

\begin{abstract}
Occlusion perception, a critical foundation for human-level spatial understanding, embodies the challenge of integrating visual recognition and reasoning. Though multimodal large language models~(MLLMs) have demonstrated remarkable capabilities, their performance on occlusion perception remains under-explored. To address this gap, we introduce O-Bench, the first visual question answering~(VQA) benchmark specifically designed for occlusion perception.
Based on SA-1B, we construct 1,365 images featuring semantically coherent occlusion scenarios through a novel layered synthesis approach. Upon this foundation, we annotate 4,588 question-answer pairs in total across five tailored tasks, employing a reliable, semi-automatic workflow.
Our extensive evaluation of 22 representative MLLMs against the human baseline reveals a significant performance gap between current MLLMs and humans, which, we find, cannot be sufficiently bridged by model scaling or thinking process. We further identify three typical failure patterns, including an overly conservative bias, a fragile gestalt prediction, and a struggle with quantitative tasks. We believe O-Bench can not only provide a vital evaluation tool for occlusion perception, but also inspire the development of MLLMs for better visual intelligence. Our benchmark will be made publicly available upon paper publication.
% comprising 4,588 question-answer pairs across five tailored tasks. O-Bench is meticulously constructed by first synthesizing realistic yet controlled, ground-truth-rich occlusion scenes, and then annotating question-answer pairs through a reliable, semi-automatic workflow.
% We first construct a set of semantically coherent occlusion scenarios based on SA-1B through a novel layered synthesis approach, which inherently provides ground-truth masks and full appearances. Employing a reliable, semi-automatic workflow, we then annotate 4,588 question-answer pairs in total across five tailored tasks.
\end{abstract}


\begin{figure}[t]
  \centering
   \includegraphics[width=0.985\columnwidth]{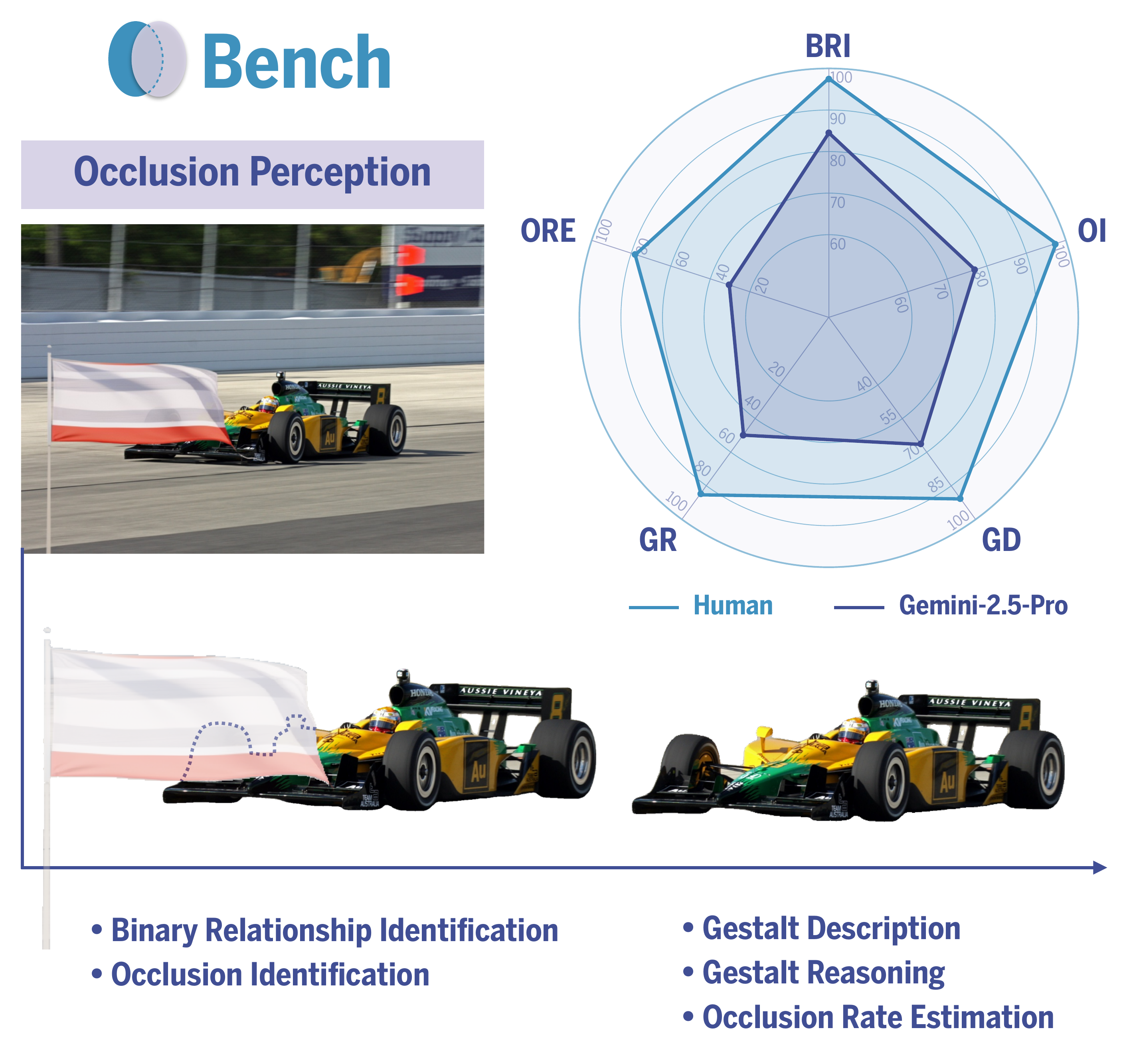}
   \caption{\textbf{O-Bench: The first occlusion perception benchmark for MLLMs.} O-Bench contains 4,588 question-answer pairs across five tailored tasks, grounded in 1,365 elaborately synthesized images. According to our evaluation, even the best-performing Gemini-2.5-Pro still has a considerable gap with human level~(upper right).}
   \label{fig:bench_intro}
\end{figure}

\begin{figure*}[t]
  \centering
   \includegraphics[width=\linewidth]{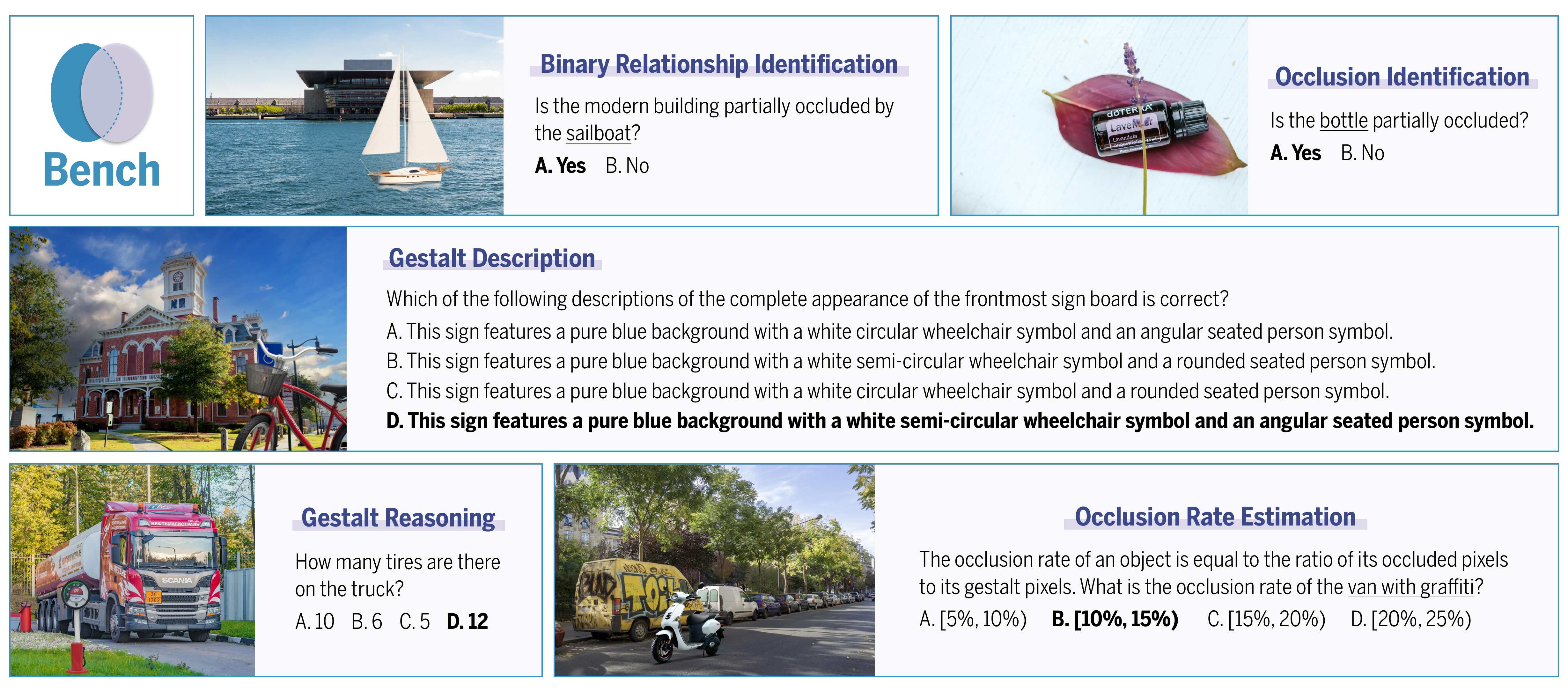}
   \caption{\textbf{Overview of O-Bench.} O-bench targets a systematic assessment for occlusion perception, comprising five tasks: Binary Relationship Identification, Occlusion Identification, Gestalt Description, Gestalt Reasoning, and Occlusion Rate Estimation. The correct option in each sample is marked in \textbf{bold}. Please zoom in for more image details.}
   \label{fig:bench_example}
\end{figure*}

\section{Introduction}
With the rapid advancement of multimodal large language models~(MLLMs), achieving human-level spatial understanding has emerged as a significant issue. Among the spatial relationships in the real world, occlusion is one of the most prevalent and fundamental phenomena~\cite{kellman1991theory}. 
For humans, perceptual completion of partially occluded objects is a cornerstone of the visual cognition, an ability present from infancy~\cite{michotte1964compléments, kellman1983perception}, enabling us to robustly interpret complex visual input and plan accordingly~\cite{ling2020variational}. 
% 对ai困难+相关研究
% 广泛应用/影响
For computer vision, similarly, occlusion perception is critical. Given the ubiquitous nature of occlusions, this ability can enhance the performance and reliability of numerous downstream tasks, such as navigation, manipulation, and scene understanding~\cite{ao2023image, ozguroglu2024pix2gestalt}.
% For computer vision models, however, the perception of partially occluded objects remains a considerable challenge, which consequently affects the performance of numerous downstream tasks, such as navigation, manipulation, and scene understanding~\cite{ao2023image}.
% ~\cite{van2024overcoming, suresh2024neuralfeels, yeh2025seeing}
Beyond the practical relevance, occlusion perception also serves as an ideal testbed for a model's ability to integrate visual recognition with reasoning, as it inherently requires identifying visible cues and inferring hidden content based on contextual evidence and prior knowledge.
% mllm是否能够
Endowed with rich prior knowledge and reasoning capabilities from large-scale training, MLLMs, correspondingly, are potentially positioned for this challenge. A natural question thus arises: 
\emph{How effectively can MLLMs tackle the challenge of occlusion perception?}

% 集合感知和推理，mllm是否能够？没有相关研究
% 提出本文测试集
However, a comprehensive study on this domain is still lacking. To address this gap, we introduce \textbf{O-Bench}, the first visual question answering~(VQA) benchmark specifically engineered for occlusion perception. 
% Actually, O-Bench is engineered not only to systematically evaluate the occlusion perception capabilities of MLLMs, but also to assess MLLMs' advanced capacity for integrating visual recognition and reasoning.
% However, the occlusion perception ability of MLLMs has remained largely unexplored. To address this gap, we propose \emph{O-Bench}, the first visual question answering~(VQA) benchmark specifically designed for occlusion perception tasks.
As shown in Fig.~\ref{fig:bench_example}, O-Bench provides a systematic evaluation suite, comprising 4,588 question-answer pairs that are distributed across five distinct tasks: \emph{(1)~Binary Relationship Identification}, \emph{(2)~Occlusion Identification}, \emph{(3)~Gestalt Description}, \emph{(4)~Gestalt Reasoning}, and \emph{(5)~Occlusion Rate Estimation}. These tasks cover a spectrum of abilities, ranging from the basic recognition of the occlusion presence to the advanced reasoning involving the occludee's complete appearance.
% O-Bench comprises over 4,500 question-answer pairs spanning five distinct tasks: (1) Relation Identification, (2) Occlusion Identification, (3) Gestalt Description, (4) Occlusion Rate Estimation, and (5) Gestalt Reasoning.
To enable such a fine-grained evaluation, existing occlusion-related datasets are not applicable due to lacking sufficient ground-truth or real-world semantic coherence. Therefore, we first novelly create 1,365 images by layered synthesis and rigorous curation, and then meticulously perform question-answer annotation through a reliable semi-automatic workflow.

Leveraging our proposed O-Bench, we conduct an extensive evaluation of 22 prominent MLLMs, including 12 open-source models as well as 10 proprietary models, and establish a human performance baseline for comparison.
% The results reveal a significant gap between the occlusion perception capabilities of current MLLMs and those of humans.
Our in-depth analysis of the experimental results yields several key findings. 
First, occlusion perception remains a considerable challenge for current MLLMs. A significant performance gap exists between MLLMs and humans, with many MLLMs achieving an accuracy merely comparable to random chance even on the basic occlusion identification task.
Second, neither model scaling nor thinking process is a sufficient strategy for mastering occlusion perception. Both approaches bring certain improvements but clearly hit a ceiling.
% Second, occlusion perception demands a robust and well-balanced model architecture. We find that the potential unlocked by deep thinking process is capped by the intrinsic strength of the base model, while the performance unleashed by a larger language model is contingent upon the capacity of its vision encoder counterpart.
% Third, we identify three typical failure patterns across most MLLMs: (1) an overly conservative judgment on occlusions, (2) a significant performance degradation at high occlusion rates, and (3) a general difficulty with quantitative reasoning.
% Third, we identify three typical failure patterns across most MLLMs: \emph{(1)~an overly conservative bias}, where MLLMs often acknowledge occluded parts only when the evidence is overwhelmingly clear; \emph{(2)~a fragile gestalt prediction} that deteriorates significantly with increasing occlusion rate; and \emph{(3)~frequent quantitative errors}, resulting in markedly lower accuracy on such questions.
Third, we identify three typical failure patterns across most MLLMs: \emph{(1)~an overly conservative bias}, where MLLMs often acknowledge occluded parts only when the evidence is overwhelmingly clear; \emph{(2)~a fragile gestalt prediction} that deteriorates significantly with increasing occlusion rate; and \emph{(3)~a struggle with quantitative tasks}.
In summary, our main contributions are threefold:
\begin{itemize}
\item We introduce O-Bench, a novel and challenging VQA benchmark for occlusion perception.
% \item We conduct an extensive evaluation of 22 representative MLLMs against a human performance baseline.
\item Extensive evaluation is conducted for 22 representative MLLMs against a human performance baseline.
\item A comprehensive analysis of the results is provided, revealing the challenges and prevalent weaknesses of MLLMs when performing occlusion perception.
\end{itemize}
We believe these contributions will inspire greater community focus on the issue of occlusion perception, and help to narrow the gap with human-level visual intelligence.

\section{Related Work}
\noindent\textbf{Multimodal Large Language Models.}
With the development of large language models~(LLMs), multimodal large language models~(MLLMs) also advance rapidly~\cite{yin2024survey}. Employing a paradigm that aligns the gap between vision and language~\cite{liu2023llava, liu2024llava1.5, li2023blip2}, recent MLLMs~\cite{wu2024deepseekvl2, bai2025qwen2.5vl, zhu2025internvl3, deitke2025molmo, hong2025glm4.1v} have demonstrated remarkable capabilities spanning numerous tasks, which progressively moving from basic perception~\cite{wu2023qbench, liu2024mmbench, li2024seed} to complex reasoning~\cite{yue2024mmmu, chen2024mmstar, lu2024mathvista} and even embodied action~\cite{majumdar2024openeqa, liu2025visualagentbench, yang2025thinking}.
Despite the impressive achievements, their performance in perceiving the real world as humans do, especially under ubiquitous conditions with occlusions, is still an open question.

\noindent\textbf{Occlusion Perception.}
Occlusion perception is a long-standing and fundamental problem in both cognitive science and computer vision. Rooted in gestalt psychology~\cite{kohler1929gestalt}, the ability to perceive a complete object from partial information is considered a cornerstone of visual intelligence~\cite{kanizsa1979organization, kellman1983perception}. In the deep learning era, prior work has extensively explored approaches for predicting the complete bounding boxes~\cite{kar2015amodal, hsieh2023tracking}, masks~\cite{li2016amodal, qi2019amodal, ke2021deep, chen2025using}, and appearances~\cite{ehsani2018segan, zhan2020self, ozguroglu2024pix2gestalt,ao2025open} of occluded instances, with applications such as autonomous driving and robotic manipulation~\cite{ao2023image}. Nevertheless, their generalization capabilities remain a significant challenge. MLLMs, with intrinsic knowledge and reasoning skills, may approach this problem from a new perspective.

\noindent\textbf{Spatial Understanding Benchmarks.}
Underpinning various real-world applications, MLLMs' spatial understanding has drawn widespread research attention. Correspondingly, a series of insightful benchmarks~\cite{liu2023vsr, kamathetal2023whatsup, wang2023po3d, shirietal2024spatialmm, wang2024spatialeval, chen2024spatialvlm, yang2025thinking, wang2025spatial457} are proposed to cover both qualitative~(\emph{e.g.}, left, right, above, below) and quantitative~(\emph{e.g.}, specific distance, clock direction) spatial relationships. However, these existing benchmarks do not comprehensively evaluate the capabilities to understand the prevalent occlusion relationship. In contrast, our O-Bench fills this critical void, providing a systematic and tailored assessment of occlusion perception in natural images.

\section{O-Bench}
We introduce O-Bench, a novel occlusion perception benchmark for MLLMs. In the following sections, we will first provide an overview of its composition and task design, and then detail its meticulous construction pipeline.
\subsection{Benchmark Overview}
\begin{figure}[t]
  \centering
   \includegraphics[width=\columnwidth]{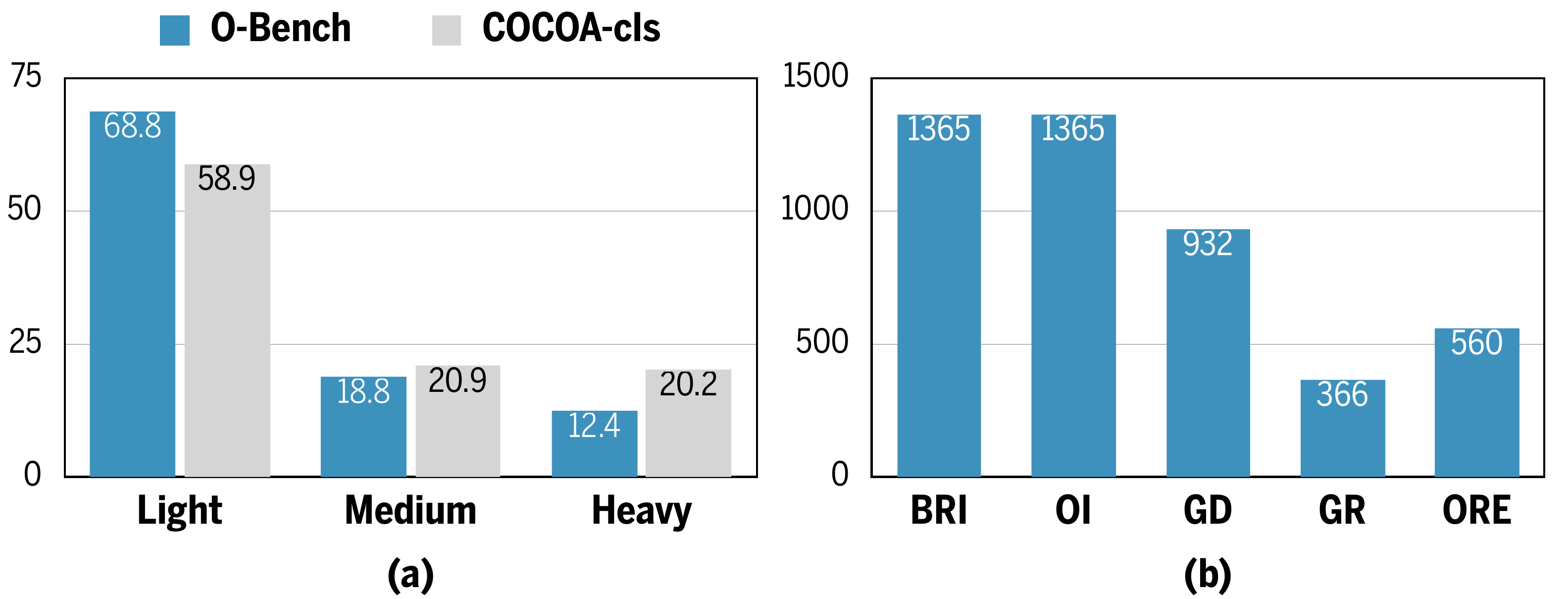}
   \caption{\textbf{Benchmark statistics.} (a) Proportions~(\%) at different occlusion levels: O-Bench vs. COCOA-cls. Light, Medium, and Heavy levels represent occlusion rates of (0, 20\%), [20\%, 40\%), and [40\%, 100\%), respectively. (b) Distribution of tasks in O-Bench.}
   \label{fig:bench_stat}
\end{figure}

% 形式化说明目标物体二元组
% TODO: object -> instance
O-Bench comprises 4,588 questions grounded in 1,365 unique images. Each image in O-Bench, denoted as $\mathbf{I}$, contains a target instance $\mathcal{T}$ and a single potential occluder $\mathcal{O}$---an object synthesized onto the foremost layer that may or may not actually occlude the target instance.
% TODO：仅考虑图片内情况，在Prompt当中包含
% To avoid subjective bias, we do not consider truncation by the image border as a form of occlusion, which is explicitly clarified in the evaluation prompts.
% TODO: 遮挡率分布图
To enhance the representativeness, the occlusion rate distribution of O-Bench mirrors that of natural images, where lower occlusion rates are more frequent. The statistics in Fig.~\ref{fig:bench_stat}(a) exhibit that O-Bench and COCOA-cls~\cite{follmann2019cocoacls}, the prominent real-world occlusion dataset, have similar proportions at each occlusion level.
% we align its occlusion rate distribution with that of natural images. This distribution typically exhibits a long-tail pattern, where lower occlusion rates are more frequent. As shown in Fig.~\ref{}, taking the COCOA-cls dataset~\cite{follmann2019cocoacls} as an example, nearly 60\% of the occluded instances are classified as Level-1 (\emph{i.e.}, with an occlusion rate under 20\%).

% VQA形式题目
% 例子放在一张图里
As shown in Fig.~\ref{fig:bench_example}, O-Bench encompasses five distinct tasks:
\textbf{(1)~Binary Relationship Identification}~(BRI) challenges the MLLM to verify a given occlusion relationship between the two instances $\mathcal{T}$ and $\mathcal{O}$, which may be factually incorrect due to a lack of overlap or an inverted foreground-background order.
\textbf{(2)~Occlusion Identification}~(OI) challenges the MLLM to determine if the target instance $\mathcal{T}$ is partially occluded without cues about the potential occluder $\mathcal{O}$, compelling it to judge whether the visible portion constitutes the instance's entirety.
\textbf{(3)~Gestalt Description}~(GD) challenges the MLLM to reason about the complete appearance of the occludee $\mathcal{T}$ and select the correct description. The answer options are intentionally designed to be similar, where the main variations lie in the descriptions of the occluded portion.
\textbf{(4)~Gestalt Reasoning}~(GR) challenges the MLLM to perform multi-step reasoning. The questions typically involve first forming a mentally completed gestalt and then leveraging it and relevant context to inform a subsequent deduction, covering both qualitative~(GR-Qual) and quantitative~(GR-Quant) subcategories.
% The GR task is divided into qualitative~(GR-Qual) and quantitative~(GR-Quant) categories.
\textbf{(5)~Occlusion Rate Estimation}~(ORE) challenges the MLLM to quantify the percentage of the occludee $\mathcal{T}$'s occluded portion. The answer options are provided as discrete bins of 5\% width~(\emph{e.g.}, $[5\%, 10\%)$), requiring the MLLM to confirm a fine-grained, pixel-level gestalt.
Fig.~\ref{fig:bench_stat}(b) shows the number of each task. All tasks are formulated as single-choice visual question answering (VQA) problems to ensure objective and convenient evaluation, among which the basic tasks~(BRI and OI) are binary-choice questions while the advanced tasks~(GD, GR, and ORE) are four-choice questions.

% TODO: 统计数据表格
% \begin{table}[b]
% \centering
% \resizebox{\columnwidth}{!}{%
% \begin{tabular}{c|c|ccccc}
% \midrule
% \makecell[c]{Occlusion\\Rate} & \makecell[c]{Unique\\Images} & BRI & OI & GD & GR & ORE \\ \midrule
% % \rowcolor{gray!15}
% % Unique Images & - & - & - & - & -  &   1365          \\
% Level-0 & 433 & 433 & 433 & - & - & -          \\
% Level-1 & 641 & 641 & 641 & 641 & 225 & 269          \\
% Level-2 & 175 & 175 & 175 & 175 & 86 & 175          \\
% Level-3 & 116 & 116 & 116 & 116 & 55 & 116          \\
% \rowcolor{gray!15}
% Total &  1365 &  1365 & 1365 & 932 & 366 & 560        \\
% \midrule
% \end{tabular}
% }
% \caption{\textbf{Statistics of O-Bench.}}
% \label{tab:stat}
% \end{table}

\begin{figure*}[t]
  \centering
   \includegraphics[width=\linewidth]{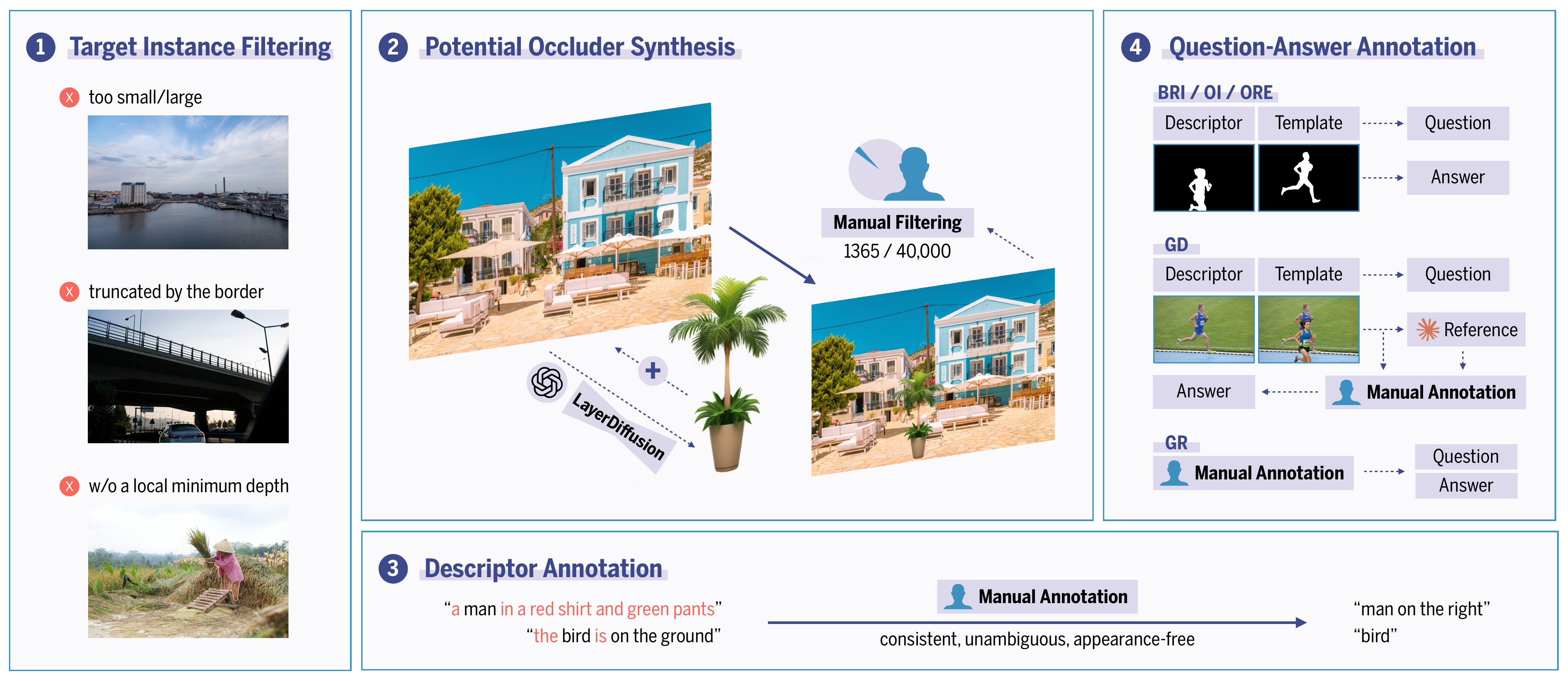}
   \caption{\textbf{Benchmark construction pipeline.} The pipeline begins with the filtering of suitable target instances from SA-1B. The potential occluder is then generated and composited into the image with a strict manual screening. Subsequently, utilizing rich known information, the descriptors and question-answer pairs are annotated through a semi-automatic workflow.}
   \label{fig:data_pipeline}
\end{figure*}

\subsection{Benchmark Construction}
\label{sec:construction}
The aforementioned evaluation of occlusion perception demands ground-truth information including both gestalt segmentation masks and appearances. While a few datasets~\cite{wang2020robust, ozguroglu2024pix2gestalt} provide such annotations, their methodology of randomly pasting foreground instances results in compositions that lack semantic plausibility and deviate from natural image characteristics. Therefore, to assess MLLMs' occlusion perception capabilities in practical scenarios, O-Bench is developed based on our meticulously created images.

As illustrated in Fig.~\ref{fig:data_pipeline}, considering the accessibility of definite hidden content, we first select natural images with suitable unoccluded target instances $\mathcal{T}$ from SA-1B~\cite{kirillov2023sam}. Then, we adopt a novel layered synthesis approach, where potential occluders $\mathcal{O}$ are intelligently suggested and independently generated, and the resulting compositions undergo a strict filtering process. Compared to employing end-to-end image editing models, this approach ensures the image quality, while meeting two critical requirements: (1)~the original image content, outside of the area occluded by $\mathcal{O}$, should remain completely unaltered, especially the visible portion of the target instance $\mathcal{T}$; and (2)~the precise segmentation mask of $\mathcal{O}$ should be automatically obtainable. Subsequently, we conduct descriptor and question-answer annotation via a reliable, semi-automatic workflow. The specific construction pipeline is detailed as follows.

\noindent\textbf{Target Instance Filtering.}
% \noindent\textbf{Image Synthesis}
% SA-1B+GranD原始数据
% DepthAnything选出无遮挡
% The foundation of O-Bench is a meticulously designed image synthesis pipeline aimed at creating controlled yet realistic occlusion scenarios. 
Our methodology commences by sourcing data from established datasets to streamline the workflow. We leverage the SA-1B dataset~\cite{kirillov2023sam} for its vast collection of natural images and corresponding segmentation masks, and augment it with text labels from GranD~\cite{rasheed2024grand}. From this pool, we curate 20,000 suitable target instances $\mathcal{T}$ by applying several filtering criteria: We first discard instances that are either too large or too small. Subsequently, using Depth Anything V2~\cite{yang2024depthanythingv2}, we select only those instances that exhibit the local minimum depth, ensuring they are not occluded in the original image. To eliminate subjective bias in the annotation stages, we also filter out any instances that are truncated by the image border.

\noindent\textbf{Potential Occluder Synthesis.}
% GPT-4o给出增添物体Prompt，LayerDiffusion生成
% 随机大小位置贴图，10选1（自动）
% 人工筛选图片
% 标准
% 结果比例
The next step is to synthetically introduce the potential occluder $\mathcal{O}$. 
% Instead of employing end-to-end image editing models, we adopt a layered synthesis approach, driven by two critical requirements that facilitate subsequent annotation: (1)~the original image content, outside of the area occluded by $\mathcal{O}$, should remain completely unaltered, especially the visible portion of the target instance $\mathcal{T}$; and (2)~the precise segmentation mask of $\mathcal{O}$ should be automatically obtainable.
% (1)~the original image, especially the visible portion of the target instance $\mathcal{T}$, should remain completely consistent; and (2)~the precise segmentation mask of the synthesized instance $\mathcal{O}$ should be automatically obtainable.
Specifically, we employ Layer Diffusion~\cite{zhang2024layerdiffusion}, a model capable of generating objects with transparent backgrounds. This model allows us to generate $\mathcal{O}$ independently, while simultaneously obtaining the corresponding segmentation mask from the alpha channel. The generation prompts are procedurally supplied by GPT-4o~\cite{hurst2024gpt4o}, which suggests a reasonable object that might appear in the given scene context. To achieve a natural and unbiased placement, each generated $\mathcal{O}$ is randomly scaled and pasted 10 times to create 10 drafts, from which GPT-4o is again utilized to select the most visually plausible composition. The above generation process is repeated twice, forming a pool of 40,000 images.
Following the automated generation, we conduct two rounds of rigorous manual filtering, yielding a final set of 1,365 images. During this process, we excluded images with deficiencies including: (1)~unnatural synthesis, such as improper sizes, unrealistic positions, or blurry edges; (2)~unintended occlusions caused by other native objects, which could be initially missed due to previous depth estimation errors; and (3)~excessive occlusion, where the target instance $\mathcal{T}$ was so heavily occluded as to be unrecognizable.

\noindent\textbf{Descriptor Annotation.}
% 目的：在问题中唯一确定目标物体 + 统一格式“名词短语” + 减少描述中的信息量
% 人工标注：尽可能简短、尽可能宽泛
To unambiguously refer to the target instance $\mathcal{T}$ and the potential occluder $\mathcal{O}$ in our VQA questions, we perform a descriptor annotation step. While the GranD dataset provides text labels for target instances $\mathcal{T}$, their direct adoption is infeasible. First, these labels exhibit significant structural inconsistencies, ranging from noun phrases (\emph{e.g.}, ``a pink dress'') to full sentences (\emph{e.g.}, ``the pot is made of metal''), with irregular article (a/an, the) usage. Second, they often contain rich appearance cues (\emph{e.g.}, ``an old, rusted truck with a missing front bumper''), which would undermine the validity of our gestalt-related tasks.
Therefore, we manually annotate descriptors for $\mathcal{T}$~(by refining GranD labels) and $\mathcal{O}$~(by authoring from scratch) based on three principles: (1)~All descriptors are normalized into noun phrases without leading articles. (2)~The descriptor should be as short as possible while ensuring it uniquely identifies the instance. (3)~The descriptor should prioritize contextual attributes (\emph{e.g.}, ``statue on the right'', ``large lantern'') over specific appearance details.
Consequently, the average length of the descriptors was shortened from 6.6 words in the original GranD labels to 2.1 words in O-Bench.

\noindent\textbf{Question-Answer Annotation.}
% 半自动标注 + 标注员介绍 + 交叉检验讨论
% BRI/OI：模板
% GD：模型生成参考选项，人工修改 -> 标准
% GR：人工挑选，人工标注
% ORE：随机删除部分+随机生成选项
% After the preceding steps, we possess a rich set of ground truth data. 
After the preceding steps, both the target instance $\mathcal{T}$ and the potential occluder $\mathcal{O}$ are equipped with their respective complete appearances, segmentation masks, and manually crafted descriptors, which enables a semi-automatic annotation pipeline for the question-answer~(QA) pairs in O-Bench. The annotation is implemented by five expert annotators~(computer science students with MLLM experience) with a cross-check process, where one annotator's work is reviewed and verified by another. The specific annotation procedure varies by task:

For BRI and OI, the QA pairs are generated in an automated manner. Questions are formed by inserting the descriptors into pre-defined templates, with the order of the two descriptors in BRI being randomly swapped to increase difficulty. The ground truth answers are then derived programmatically by checking for an overlap between the segmentation masks of $\mathcal{T}$ and $\mathcal{O}$. For BRI, however, an additional check on the descriptor order is also incorporated.
% such as ``Is the [descriptor of $\mathcal{T}$] partially occluded?'' for OI, and ``Is the [descriptor of $\mathcal{T}$] partially occluded by the [descriptor of $\mathcal{O}$]?'' for BRI.

For GD, we employ a model-assisted approach. Claude-3.5-Sonnet is first utilized to generate descriptions for the full appearance of $\mathcal{T}$, which then inspire annotators to manually author a precise description to serve as the correct option. Subsequently, three distinct incorrect options are created by introducing subtle modifications to this correct description, primarily concerning the occluded portion of $\mathcal{T}$.

For GR, annotators first select images featuring rich context, and then compose questions on a case-by-case basis. These questions, containing a mix of 201 qualitative and 165 quantitative questions, are designed to require both the completed gestalt of $\mathcal{T}$ and the context information, thus typically involving multi-step reasoning.

For ORE, the ground-truth occlusion rate is directly determined based on the known segmentation masks of $\mathcal{T}$ and $\mathcal{O}$. A strategic sampling is performed to prevent MLLMs from exploiting the statistical prior that low-occlusion instances are more prevalent. We randomly retained only one-third of the images with occlusion rates in the interval $(0\%, 15\%)$. For each question, the options consist of four consecutive, non-overlapping 5\%-wide bins~(\emph{e.g.}, $[5\%, 10\%)$, $[10\%, 15\%)$, $[15\%, 20\%)$, $[20\%, 25\%)$). The discretization of the answer space not only unifies the task within the multiple-choice framework, but more importantly, it improves the evaluation robustness and aligns the task to a human-achievable level.

\begin{table*}[t]
\centering
\resizebox{0.985\linewidth}{!}{%
\begin{tabular}{c|l|c|ccccc|ccc}
\midrule
% \multirow{2}{*}{Model} & \multirow{2}{*}{Size} & \multicolumn{5}{c|}{Accuracy} & \multirow{2}{*}{O-Score} \\ \cmidrule{3-7}
%  & & BRI & OI & GD & GR & ORE & \\ \midrule
 % & Model &
\multicolumn{2}{l|}{Model} & 
$\ \ $Size$\ \ $ & $\ \ \ $BRI$\ \ \ $ & $\ \ \ $OI$\ \ \ $ & $\ \ \ $GD$\ \ \ $ & $\ \ \ $GR$\ \ \ $ & $\ \ \ $ORE$\ \ \ $ & $\ $Basic$\ $ & Advanced & \textbf{O-Score} \\ \midrule
\rowcolor{gray!15}
\multicolumn{11}{c}{Small-size MLLMs~($\sim$10B)} \\ \midrule
1 & Qwen2.5-VL & 7B & 65.3 & 45.8 & 46.8 & 26.2 & \textbf{27.5} & 55.6 & 33.5 & 42.3          \\
2 & InternVL3 & 8B & 62.6 & 49.9 & 52.5 & 33.9 & 24.8 & 56.3 & 37.0 & 44.7          \\
3 & GLM-4.1V & 9B & \textbf{68.9} & 47.1 & \textbf{53.0} & 36.1 & 25.5 & 58.0 & \textbf{38.2} & 46.1          \\
4 & Molmo & 7B & 56.8 & \textbf{64.5} & 48.4 & \textbf{43.7} & 22.3 & \textbf{60.7} & 38.1 & \textbf{47.1}          \\
\midrule
\rowcolor{gray!15}
\multicolumn{11}{c}{Medium-size MLLMs~(10B$\sim$40B)} \\ \midrule
5 & DeepSeek-VL2 & 16B & 46.0 & 42.1 & 39.6 & 19.4 & 18.0 & 44.1 & 25.7 & 33.0          \\
6 & DeepSeek-VL2 & 27B & 57.9 & 38.5 & 40.6 & 31.4 & \textbf{32.0} & 48.2 & 34.7 & 40.1          \\
7 & InternVL3 & 14B & 66.7 & 46.0 & 54.3 & 35.3 & 27.5 & 56.4 & 39.0 & 46.0          \\
8 & Qwen2.5-VL & 32B & 71.5 & \textbf{56.7} & 47.5 & 38.3 & 27.3 & \textbf{64.1} & 37.7 & 48.3          \\
9 & InternVL3 & 38B & \textbf{73.2} & 53.4 & \textbf{59.3} & \textbf{41.5} & 28.4 & 63.3 & \textbf{43.1} & \textbf{51.2}          \\
\midrule
\rowcolor{gray!15}
\multicolumn{11}{c}{Large-size MLLMs~(40B$\sim$)} \\ \midrule
10 & Qwen2.5-VL & 72B & \textbf{71.8} & 51.6 & 51.6 & 36.3 & 24.1 & 61.7 & 37.3 & 47.1          \\
11 & InternVL3 & 78B & 70.7 & 55.8 & \textbf{58.7} & 40.4 & \textbf{29.3} & 63.3 & \textbf{42.8} & 51.0          \\
12 & Molmo & 72B & 63.2 & \textbf{68.6} & 54.4 & \textbf{45.1} & 26.8 & \textbf{65.9} & 42.1 & \textbf{51.6}          \\
\midrule
\rowcolor{gray!15}
\multicolumn{11}{c}{Proprietary MLLMs} \\ \midrule
13 & Doubao-1.5-Vision-Pro$\ \ $ & - & 73.1 & 42.9 & 44.7 & 26.0 & 17.1 & 58.0 & 29.3 & 40.8          \\
14 & Claude-3.7-Sonnet & - & 67.0 & 52.6 & 51.3 & 30.1 & 28.8 & 59.8 & 36.7 & 46.0          \\
15 & Doubao-Seed-1.6 & - & 73.1 & 57.0 & 52.4 & 33.9 & 25.9 & 65.1 & 37.4 & 48.5          \\
16 & Gemini-2.0-Flash & - & 52.5 & 70.9 & 53.3 & 38.5 & 30.9 & 61.7 & 40.9 & 49.2          \\
17 & Claude-4-Sonnet & - & 73.4 & 55.1 & 56.2 & 38.8 & 27.9 & 64.3 & 41.0 & 50.3          \\
18 & Claude-4-Opus & - & 72.2 & 51.4 & 60.1 & 41.8 & 30.4 & 61.8 & 44.1 & 51.2          \\
19 & GPT-4o & - & 76.8 & 72.6 & 53.5 & 34.2 & 26.3 & 74.7 & 38.0 & 52.7          \\
20 & GPT-4.1 & - & 73.1 & 75.0 & 51.1 & 37.4 & 32.1 & 74.1 & 40.2 & 53.7          \\
21 & Gemini-2.5-Flash & - & 82.1 & \textbf{76.9} & 59.2 & 48.9 & 30.9 & 79.5 & 46.3 & 59.6          \\
22 & Gemini-2.5-Pro & - & \textbf{84.5} & 76.7 & \textbf{66.3} & \textbf{51.1} & \textbf{32.5} & \textbf{80.6} & \textbf{50.0} & \textbf{62.2}          \\
\midrule
\rowcolor{gray!15}
\multicolumn{11}{c}{Baselines} \\ \midrule
- & Random & - & 50.0 & 50.0 & 25.0 & 25.0 & 25.0 & 50.0 & 25.0 & 35.0          \\
- & Human* & - & 97.0 & 96.7 & 91.3 & 84.0 & 79.0 & 96.9 & 84.8 & 89.6         \\
% Gemini-2.5-Pro* & - & 83.3 & 76.0 & 67.7 & 51.3 & 33.3 & 79.7 & 50.8 & 62.3         \\
\midrule
\end{tabular}
}
\caption{\textbf{Evaluation results of 22 MLLMs on O-Bench.} All scores are reported as accuracy~(\%). O-Score is the mean of all five task accuracies, while the mean accuracies of basic~(BRI and OI) and advanced~(GD, GR, and ORE) tasks are also presented. The best performance in each category is marked in \textbf{bold}. *: Measured on a subset of 1,500 questions.}
\label{tab:perf}
\end{table*}

\section{Experiments}
\subsection{Evaluation Setup}
% 模型
\noindent\textbf{Model Selection.}
To provide a comprehensive and up-to-date assessment, our evaluation encompasses a diverse set of 22 recent MLLMs. This broad selection consists of two groups: 12 representative open-source models, including Qwen2.5-VL~\cite{bai2025qwen2.5vl}, InternVL3~\cite{zhu2025internvl3}, GLM-4.1V~\cite{hong2025glm4.1v}, Molmo~\cite{deitke2025molmo}, and DeepSeek-VL2~\cite{wu2024deepseekvl2} series; as well as 10 prominent proprietary models, including GPT~(GPT-4.1, GPT-4o), Gemini~(Gemini-2.5-Pro, Gemini-2.5-Flash, Gemini-2.0-Flash), Claude~(Claude-4-Opus, Claude-4-Sonnet, Claude-3.7-Sonnet), and Doubao~(Doubao-Seed-1.6, Doubao-1.5-Vision-Pro) families.

% 人类测试
\noindent\textbf{Human Baseline.}
Aiming to establish a human baseline on O-Bench, we recruit three volunteers to participate in the test. To mitigate potential performance degradation due to participant fatigue while enhancing question coverage, we create three distinct, non-overlapping subsets of 500 questions, each randomly sampled with 100 questions per task, and assign one set to each volunteer.
% ~(see Appendix for a validation of the representativeness). 
The final human baseline is calculated by averaging the performance of the three volunteers, mitigating the influence of individual biases.

% TODO: Metric
\noindent\textbf{Metric.}
The performance on each task is naturally measured by accuracy. We define the overall O-Score $S_o$ as the arithmetic mean of individual accuracies across the five tasks: $$S_o=\frac{1}{|\mathcal{Q}|}\sum_{q\in\mathcal{Q}}\bar{A}_q,$$ where $\mathcal{Q}=\{\text{BRI, OI, GD, GR, ORE}\}$ is the set of tasks and $\bar{A}_q$ denotes the accuracy on a specific task $q\in\mathcal{Q}$. O-Score ensures that each task contributes equally, regardless of the number of questions it contains. Similarly, we also report the average accuracy for the two basic tasks~(BRI and OI) and the three advanced tasks~(GD, GR, and ORE) separately, providing an auxiliary view of an MLLM's performance across different levels of difficulty.

\subsection{Results and Analysis}
The evaluation results of the 22 MLLMs on O-Bench is reported in Table~\ref{tab:perf}. From these experiments, we conduct an in-depth analysis and distill several key findings.

% 与人类水平对比
\begin{tcolorbox}
[colback=gray!10,colframe=black,boxrule=0.5pt]
\emph{\textbf{Finding 1:} Occlusion perception remains a considerable challenge for MLLMs, with a significant performance gap observed compared to humans.}
\end{tcolorbox}

% TODO：修改与人类水平相关的数据
As presented in Table~\ref{tab:perf}, even the top-performing MLLM, Gemini-2.5-Pro~\cite{comanici2025gemini}, lags behind the human baseline by a remarkable 27.4 points on O-Score, indicating a general disparity between MLLMs and humans in occlusion perception.
On the \emph{basic} tasks, leading proprietary MLLMs like the Gemini-2.5 series demonstrate substantial competence, achieving accuracies around 80\%, yet these still fall notably short of the near-perfect human proficiency. Interestingly, on the relatively straightforward Occlusion Identification~(OI) task, the performance of many MLLMs hovers near the 50\% random-chance level, which suggests that even the basic recognition of an occlusion's presence is not yet a solved problem for MLLMs.
When moving to the \emph{advanced} tasks, the performance gap with humans widens into a veritable chasm. The highest score achieved by evaluated MLLMs is a mere 50.0, which is dwarfed by the human baseline of 84.8, and is even closer to the random-chance level~(25.0) than to human performance. Meanwhile, the performance difference between proprietary and open-source MLLMs narrows since all models struggle universally. This widespread failure indicates that MLLMs fundamentally lack the robust gestalt completion capabilities that are cornerstones of human visual perception.

% 关于scaling和thinking
\begin{tcolorbox}
[colback=gray!10,colframe=black,boxrule=0.5pt]
\emph{\textbf{Finding 2:} Neither scaling nor ``thinking'' is a sufficient strategy for mastering occlusion perception.}
\end{tcolorbox}

% TODO
First, performance gains are not guaranteed by simply scaling the model. As shown in Table~\ref{tab:perf}, scaling up can offer initial improvements. For instance, InternVL3-38B achieves a 5.2-point higher O-Score than InternVL3-14B. According to the report\cite{zhu2025internvl3}, this substantial gain coincides with an upgrade in both its language model~(from 14B to 32B) and its vision encoder~(from 300M to 6B). However, a clear bottleneck emerges at larger scales, as evidenced by the performance stagnation of InternVL3~(from -38B to -78B) and Qwen2.5-VL~(from -32B to -72B) series. We speculate this performance ceiling is reached when the demand for rich and nuanced visual features exceeds the vision encoder's capacity, which calls for a more powerful vision backbone trained on a wider variety of data.

Second, we explore the effect of the ``thinking'' process. Leveraging a setting supported by Gemini-2.5-Flash, Claude-4-Opus, and Doubao-Seed-1.6 to disable deep thinking, we conduct an ablation study. As shown in Table~\ref{tab:thinking}, enabling the thinking process provides a notable performance boost: The O-Score of Gemini-2.5-Flash improves by 5.1 points. Claude-4-Opus and Doubao-Seed-1.6 also see increases of 5.2 and 5.3 points, respectively. However, the thinking-enhanced version of Claude-4-Opus~(51.2) and Doubao-Seed-1.6~(48.5) still performs significantly worse than the non-thinking version of Gemini-2.5-Flash~(54.5), which indicates that the effectiveness of the thinking process is fundamentally predicated on the inherent capabilities of the base model. Therefore, true advancement in occlusion perception likely requires a synergistic improvement of both the foundation model and the reasoning strategies.
% However, the improvement is ultimately constrained by the model's inherent capabilities. The thinking-enhanced version of Claude-4-Opus~(51.2) and Doubao-Seed-1.6~(48.5) still performs significantly worse than the non-thinking version of Gemini-2.5-Flash~(54.5), which indicates that the effectiveness of the thinking process is fundamentally predicated on the strength of the base model.

\begin{table}[t]
\centering
\resizebox{\columnwidth}{!}{%
\begin{tabular}{l|ccccc|c}
\midrule
Model &  BRI & OI & GD & GR & ORE & \textbf{O-Score} \\
\midrule
\rowcolor{gray!15}
\multicolumn{7}{c}{w/o thinking process} \\ \midrule
Doubao-Seed-1.6 & 73.5 & 49.5 & 46.9 & 25.7 & 20.2 & 43.2          \\
Claude-4-Opus & 69.9 & 43.5 & 55.4 & 34.4 & 26.8 & 46.0          \\
Gemini-2.5-Flash & 72.6 & 76.6 & 53.8 & 42.6 & 27.1 & 54.5          \\
\midrule
\rowcolor{gray!15}
\multicolumn{7}{c}{w/ thinking process} \\ \midrule
Doubao-Seed-1.6 & 73.1 & 57.0 & 52.4 & 33.9 & 25.9 & 48.5          \\
Claude-4-Opus & 72.2 & 51.4 & 60.1 & 41.8 & 30.4 & 51.2          \\
Gemini-2.5-Flash & 82.1 & 76.9 & 59.2 & 48.9 & 30.9 & 59.6          \\
\midrule
\end{tabular}
}
\caption{\textbf{Ablation study on the effect of the thinking process.} We compare the performance of three representative MLLMs with and without a ``thinking'' process.}
\label{tab:thinking}
\end{table}

\begin{figure}[t]
  \centering
   \includegraphics[width=\columnwidth]{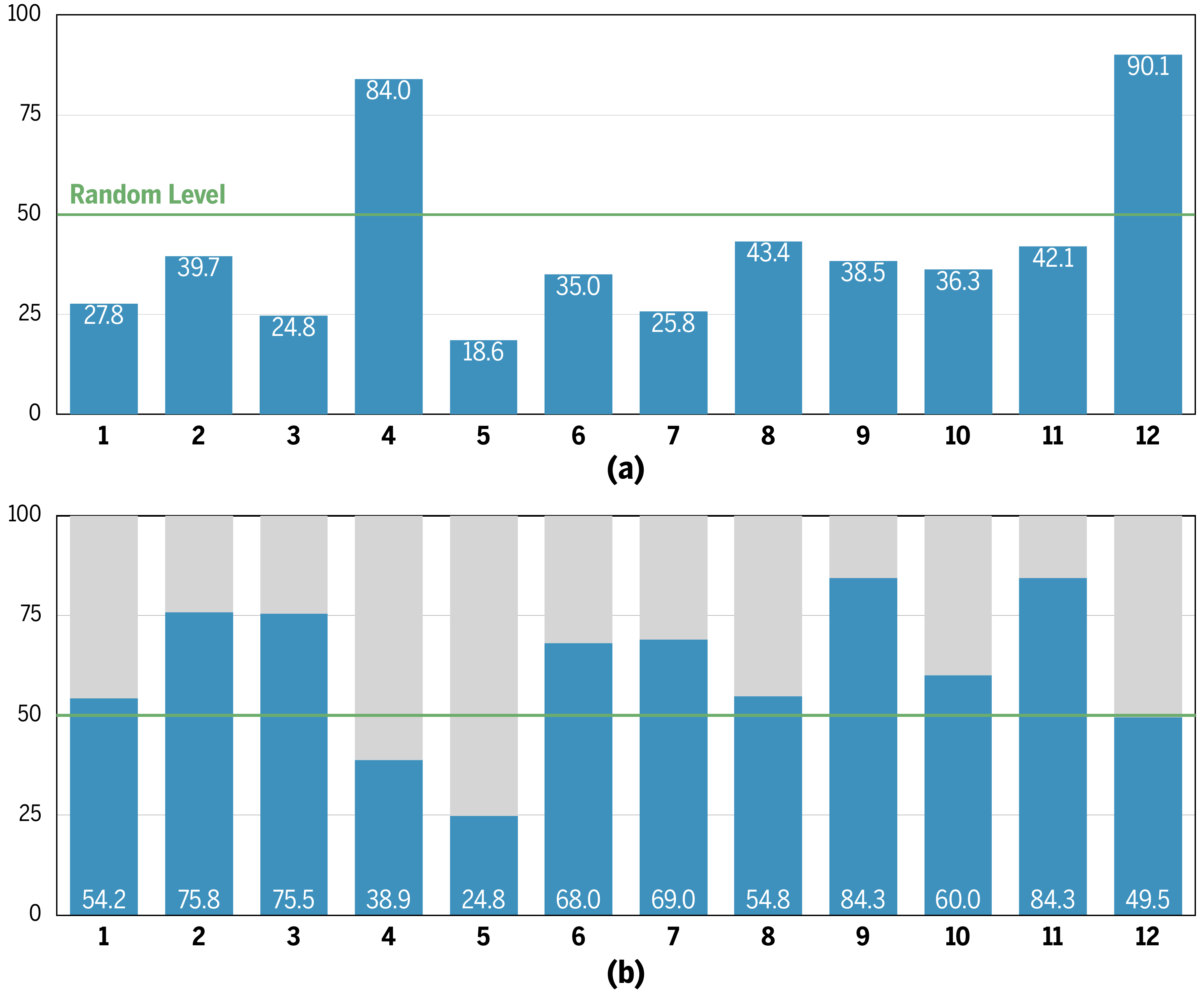}
   \caption{\textbf{Overly conservative bias.} (a)~Recall~(\%) on the OI task, where most MLLMs are below the random level~(green line). (b)~Error distribution~(\%) on the ORE task. The bars show the proportion of underestimation~(blue) versus overestimation~(gray) within all errors for each model. In this figure, x-axis denotes the model index as shown in Table~\ref{tab:perf}.}
   % MLLMs are labeled with the same index in Table~\ref{tab:perf}.}
   \label{fig:conservative}
\end{figure}
% The proportions~(\%) of ORE errors

% failure patterns: 保守、遮挡率、定量
% TODO: a difficulty in quantitative questions?
\begin{tcolorbox}
[colback=gray!10,colframe=black,boxrule=0.5pt]
\emph{\textbf{Finding 3:} Typical failure patterns include an overly conservative bias, a fragile gestalt prediction, and a struggle with quantitative tasks.}
\end{tcolorbox}

% TODO: 画图
\emph{Overly Conservative Bias.} MLLMs possess a pronounced tendency towards excessive caution. As illustrated in Fig.~\ref{fig:conservative}(a), MLLMs demonstrate a systematically low recall rate on the OI task. A majority of these open-source MLLMs even exhibit a recall below the 50\% random level. Many MLLMs default to a ``safe'' negative response, preferring to miss a true occlusion rather than risk a false positive. For instance, half of evaluated MLLMs mistakenly choose ``No'' for the OI question shown in Fig.~\ref{fig:bench_example}. This bias can also be confirmed in the ORE task. As shown in Fig.~\ref{fig:conservative}(b), when MLLMs make an error in this task, they are significantly more likely to underestimate the occlusion rate than to overestimate it. For these open-source MLLMs, underestimations constitute 61.1\% of all errors on average.

\begin{figure}[t]
  \centering
   \includegraphics[width=\columnwidth]{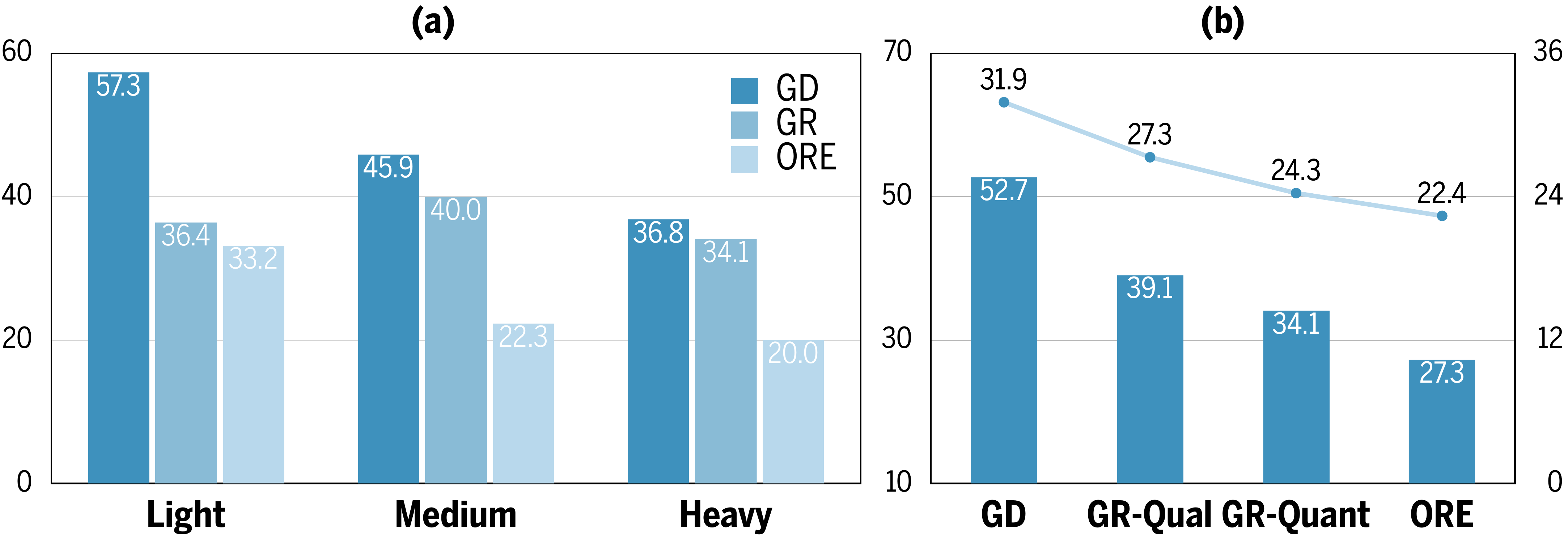}
   \caption{\textbf{Analysis of occlusion levels and task types.} (a)~Average model-wise accuracy~(\%) across different occlusion levels. The Light, Medium, and Heavy levels correspond to occlusion rates of (0, 20\%), [20\%, 40\%), [40\%, 100\%), respectively. (b)~Performance comparison across different tasks. The bars and the line represent average model-wise accuracy~(\%) and standard deviation~(\%), respectively.}
   \label{fig:task_compare}
\end{figure}

% TODO: 画图
\emph{Fragile Gestalt Prediction.} We find that the gestalt completion capabilities of MLLMs are sensitive to the degree of occlusion. 
A stark illustration is the GD example in Fig.~\ref{fig:bench_example}~(44\% occlusion rate), which only 1/22 MLLM answers correctly.
To quantify this, we introduce the average model-wise accuracy $\bar{A}^{m}$. For a question set $q$ and $N$ evaluated models, $\bar{A}_q^{m}=\frac{1}{|q|}\sum_{j\in q}A_j^m=\frac{1}{N|q|}\sum_{i=1}^N\sum_{j\in q}s_{i,j}$, where $s_{i,j}$ is the binary score for model $i$ on question $j$.
We measure the average model-wise accuracy across three occlusion levels~(Light, Medium, and Heavy). As shown in Fig.~\ref{fig:task_compare}(a), the performance degradation is evident in the GD task. The average model-wise accuracy drops by over 20\%, from 57.3\%~(12.6/22) on the Light subset to 36.8\%~(8.1/22) on the Heavy subset. Similarly, this trend can also be observed in other gestalt-related advanced tasks. This consistent fragility indicates that current MLLMs' gestalt reasoning heavily relies on sufficient visible cues and struggles to cope with more complex, information-sparse scenarios.

% TODO: 画图
\emph{Struggle with Quantitative Tasks.}
Across the board, tasks requiring numerical or proportional judgment are substantially more challenging for MLLMs. This disparity is starkly illustrated by a cross-task comparison of average model-wise accuracies $\bar{A}^{m}$. As shown in Fig.~\ref{fig:task_compare}(b), the ORE task, which is purely quantitative, stands out as the most difficult challenge in the entire benchmark, with a mere 27.3\% average accuracy. Within the GR task, the average accuracy on quantitative~(GR-Quant) subcategory~(34.1\%) is notably lower than on qualitative~(GR-Qual) subcategory~(39.1\%). For example, only 2/22 evaluated MLLMs correctly answer the quantitative GR question in Fig.~\ref{fig:bench_example}.
Moreover, for quantitative tasks, the standard deviation of the model-wise accuracies $A^m$ is also relatively low, which further confirms that quantitative reasoning is a universal bottleneck.

\section{Conclusion}
In this paper, we introduce O-Bench, a novel and challenging benchmark specifically designed to evaluate the occlusion perception capabilities of MLLMs. Through a meticulous data construction pipeline and a systematic task design, O-Bench provides an extensive assessment for 22 leading MLLMs. We reveal a significant performance gap between current MLLMs and humans, and find that neither scaling nor thinking process is sufficient to overcome this challenge. Furthermore, we categorize typical failure patterns, including an overly conservative bias, a fragile gestalt prediction, and a struggle with quantitative tasks. We believe that O-Bench will not only serve as an evaluation tool but also as a foundation for related work, inspiring the development of MLLMs with human-level visual intelligence.

\bibliography{aaai2026}

\end{document}